\documentclass[runningheads]{llncs}
\usepackage{graphicx}
\usepackage{amsmath,amssymb} 
\usepackage{color}
\usepackage{epsfig}
\usepackage{graphicx}
\usepackage{amsmath}
\usepackage{amssymb}
\usepackage{chngcntr}
\usepackage{algorithm}
\usepackage[noend]{algpseudocode}
\usepackage{xcolor}
\usepackage{paralist}
\usepackage[toc,page]{appendix}
\usepackage{wrapfig}
\usepackage{changepage}

\usepackage[symbol]{footmisc}

\usepackage{color}
\usepackage[utf8]{inputenc} 
\usepackage[T1]{fontenc}    
\usepackage[pagebackref=true,breaklinks=true,colorlinks,bookmarks=false,citecolor=blue]{hyperref} 
\usepackage{url}            
\usepackage{amsfonts}       
\usepackage{nicefrac}       
\usepackage{microtype}      
\usepackage{subcaption}
\usepackage[font=footnotesize,labelfont=footnotesize]{caption}
\usepackage{multirow}
\usepackage[toc,page]{appendix}
\usepackage{mysymbols}
\usepackage{cleveref}
\usepackage{float}
\usepackage{booktabs}       
\usepackage{comment}

\newcommand{\mohamed}[1]{\textcolor{green}}
\newcommand{\sml}[1]{\textcolor{purple}}
\newcommand{\devi}[1]{\textcolor{red}}
\newcommand{\pc}[1]{\textcolor{blue}}
\newcommand{\rp}[1]{\textcolor{red}}

\begin{document}
\title{Choose Your Neuron: Incorporating Domain Knowledge through Neuron-Importance} 

\titlerunning{Choose Your Neuron}
%


\author{Ramprasaath R. Selvaraju\inst{1}$^\dagger$\thanks{Equal Contribution} \and
Prithvijit Chattopadhyay\inst{1*} \and \\
Mohamed Elhoseiny\inst{2} \and 
Tilak Sharma\inst{2} \and
Dhruv Batra\inst{1,2} \and \\
Devi Parikh\inst{1,2} \and
Stefan Lee\inst{1}}

%
\authorrunning{Ramprasaath R. Selvaraju and P. Chattopadhyay}


%
\institute{$^1$Georgia Institute of Technology \quad $^2$Facebook \\
\email{\{ramprs,prithvijit3,dbatra,parikh,steflee\}@gatech.edu}\\
\email{\{elhoseiny,tilaksharma,dbatra,parikh\}@fb.com}}
\maketitle              
%

\begin{abstract}


Individual neurons in convolutional neural networks supervised for image-level 
classification tasks have been shown to implicitly learn semantically meaningful concepts ranging from simple textures 
and shapes to whole or partial objects -- forming a ``dictionary'' of concepts 
acquired through the learning process.
%
In this work we introduce a simple, efficient zero-shot learning approach based on this observation. Our approach, which we call Neuron Importance-Aware Weight Transfer (NIWT),  learns to map domain knowledge about novel ``\emph{unseen}'' classes onto this dictionary of learned concepts 
and then optimizes for network parameters that can effectively combine these concepts -- essentially 
learning classifiers by discovering and composing learned semantic concepts in deep networks.
%
{Our approach shows improvements over previous approaches on the CUBirds and AWA2 generalized zero-shot learning benchmarks.}
We demonstrate our approach on a diverse set of semantic inputs as external domain knowledge including attributes and natural language captions. 
Moreover by learning inverse mappings, NIWT can provide visual and textual explanations for the predictions made by the newly learned classifiers and provide neuron names. 
Our code is available at {\small \url{https://github.com/ramprs/neuron-importance-zsl}}.
\keywords{Zero Shot Learning \and Interpretability \and Grad-CAM}
\end{abstract}


\section{Introduction}
\footnote[2]{Work done partly at Facebook} Deep neural networks have pushed the boundaries of standard classification tasks in the past few years, with performance on many challenging benchmarks reaching near human-level accuracies. 
One caveat however is that these deep models require massive labeled datasets -- failing to generalize from few examples or descriptions of unseen classes like humans can.
To close this gap, the task of learning deep classifiers for unseen classes from external domain knowledge alone -- termed zero-shot learning (ZSL) -- has been the topic of increased interest within the community \cite{larochelle2008zero,lampert2014attribute,farhadi2009describing,norouzi2013zero,socher2013zero,zhang2015zero,xian2016latent,akata2016label,frome2013devise,akata2015evaluation,romera2015embarrassingly,changpinyo2016synthesized,kodirov2017semantic}.

As humans, much of the way we acquire and transfer knowledge about novel concepts is in reference to or via composition of concepts which are already known. For instance, upon hearing that \emph{``A Red Bellied Woodpecker is a small, round bird with a white breast, red crown, and spotted wings.''}, we can compose our understanding of colors and birds to imagine how we might distinguish such an animal from other birds. However, applying a similar compositional learning strategy for deep neural networks has proven challenging. 

\begin{figure}[t]
	\centering
	\includegraphics[width=\textwidth]{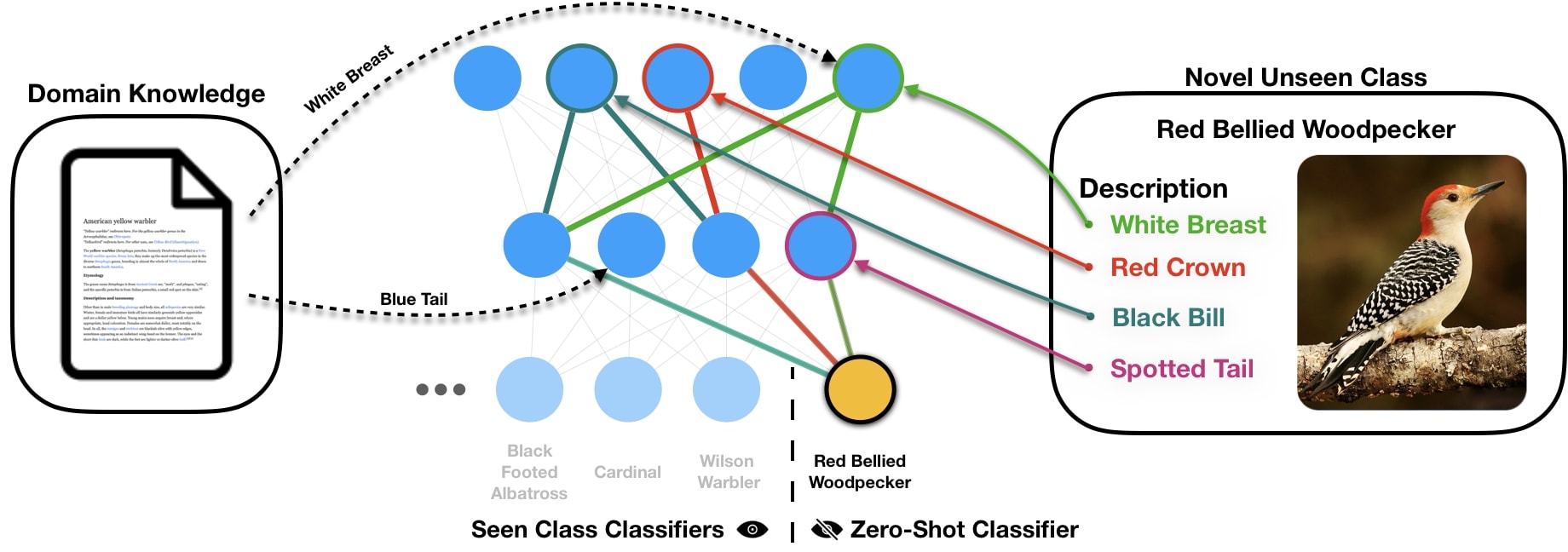}
	\caption{We present our Neuron Importance-Aware Weight Transfer (NIWT) approach which maps free-form domain knowledge about unseen classes to relevant concept-sensitive neurons within a pretrained deep network.  We then optimize the weights of a novel classifier such that the activation of this set of neurons results in high output scores for the unseen classes in the generalized zero-shot learning setting.\\[-30pt]}
	\label{fig:teaser}
\end{figure}

While individual neurons in deep networks have been shown to learn localized, semantic concepts, these units lack referable groundings -- \ie even if a network contains units sensitive to \emph{``white breast''} and \emph{``red crown''}, there is no explicit mapping of these neurons to the relevant language name or description. This observation encouraged prior work in interpretability to crowd-source ``neuron names'' to discover these groundings \cite{netdissect2017}.  However, this annotation process is model dependent and needs to be re-executed for each model trained, which 
makes it
expensive and impractical. 
Moreover, even if given perfect 
``neuron names'', it is an open question how to leverage this neuron-level descriptive supervision to train novel classifiers. This question is at the heart of our approach.

Many existing zero-shot learning approaches make use of deep features (\ie vectors of activations from some late layer in a network pretrained on some large-scale task) to learn joint embeddings with class descriptions ~\cite{Xian_2017_CVPR,akata2013label,akata2015evaluation,changpinyo2016synthesized,qiao2016less,elhoseiny2013write,Elhoseiny_2017_CVPR,Elhoseiny_Write_2016}. 
{These higher-level features collapse many underlying concepts in the pursuit of class discrimination; consequentially, accessing lower-level concepts and recombining them in new ways to represent novel classes is difficult with these features.}
{Mapping class descriptions to lower-level activations directly on the other hand is complicated by the high 
intra-class
variance of activations due to both spatial and visual differences within instances of a class.} 
%
Our goal is to address these challenges by grounding class descriptions (including attributes and free-form text) to the \emph{importance} of lower-layer neurons to final network decisions \cite{gradcam_arxiv}.

In our approach, which we call Neuron Importance-based Weight Transfer (NIWT), we learn a mapping between class-specific domain knowledge and the importances of individual neurons within a deep network. This mapping is learnt using images (to compute neuron-importance) and corresponding domain knowledge representation(s) of training classes. 
We then use this learned mapping to predict neuron importances from knowledge about unseen classes and optimize classification weights such that the resulting network aligns with the predicted importances.
In other words, based on domain-knowledge of the unseen categories, we can predict which low-level neurons should matter in the final classification decision. 
 We can then learn network weights such that the neurons predicted to matter actually do contribute to the final decision.
In this way, we connect the description of a previous unseen category to weights of a classifier that can predict this category at test time -- all without having seen a single image from this category.
{To the best of our knowledge, this is the first zero-shot learning approach to align domain knowledge to intermediate neurons within a deep network.}
As an additional benefit, the learned mapping from domain knowledge to neuron importances grounds the neurons in interpretable semantics; automatically performing neuron naming.

We focus on the challenging generalized zero-shot (GZSL) learning setting. {Unlike standard ZSL settings which evaluate performance only on unseen classes, GZSL considers both unseen and seen classes to measure the performance. In effect, GZSL is made more challenging by dropping the unrealistic assumption that test instances are known a priori to be from unseen classes in standard ZSL.} We validate our approach across two standard datasets - Caltech-UCSD Birds (CUB) \cite{WahCUB_200_2011} and Animals with Attributes 2 (AWA2) \cite{Xian_2017_CVPR} - showing improved performance over existing methods. 
Moreover, we examine the quality of our grounded explanations for classifier decisions through textual and visual examples. \\[-8pt]

\noindent
\textbf{Contributions.} Concretely, we make the following contributions in this work:
\begin{compactitem}[$\circ$]
%
%
\item We introduce a zero-short learning approach based on mapping unseen class descriptions to neuron importance within a deep network and then optimizing unseen classifier weights to effectively combine these concepts. We demonstrate the effectiveness of our approach by reporting 
improvements on the
generalized zero-shot 
benchmark
on CUB and AWA2. We also show our approach can handle arbitrary forms of domain knowledge including attributes and captions. \\[-10pt] 
\item In contrast to existing approaches, our method is capable of explaining its zero-shot predictions with human-interpretable semantics from attributes. We show how inverse mappings from neuron importance to domain knowledge can also be learned to provide interpretable visual and textual explanations for the decisions made by newly learned classifiers for seen and unseen classes.\\[-20pt]
\end{compactitem}

\section{Related Work}

\noindent 
\textbf{Model Interpretability.}~ 
Our method aligns human interpretable domain knowledge to neurons within deep neural 
networks, instilling these neurons with understandable semantic meanings. 
There has been significant recent interest in building machine learning models that are 
transparent and interpretable in their decision making process.
For deep networks, several works propose explanations 
based on internal states or structures of the network \cite{zeiler_eccv14,GoyalMPB16,Zhou2014ObjectDE,gradcam_arxiv}. Most related to our work is the approach of Selvaraju \etal\cite{gradcam_arxiv} which computes neuron importance as part of a visual explanation pipeline. In this work, we leverage these importance scores to embed free-form domain knowledge to individual neurons in a deep network and train new classifiers based on this information. In contrast, Grad-CAM~\cite{gradcam_arxiv} simply visualizes the importance of input regions.\\[-10pt] 

\par \noindent
\textbf{Attribute-based Zero-Shot Learning.}~ 
One long-pursued approach for zero-shot learning is to leverage knowledge about common attributes
and shared parts (e.g.,  furry,
in addition to being simpler and more efficient
~\cite{romera2015embarrassingly,akata2015evaluation,akata2016label,Xian_2017_CVPR}. \\[-10pt]

\par \noindent
\textbf{Text-Based Zero-Shot Learning (ZSL).}~
In parallel research, pure text articles extracted from the web have been leveraged instead  of attributes to design zero-shot visual classifiers ~\cite{elhoseiny2013write}. 
The description of a new category is purely textual (avoiding the use of attributes) and could be extracted easily by just mining article(s) about the class of interest from the web (e.g., Wikipedia). 
Recent approaches have adopted deep neural network based classifiers, leading to a noticeable improvement on zero-shot accuracy (Bo \etal~\cite{lei2015predicting}).
The proposed approaches mainly rely on learning a similarity function between text descriptions and images (either linearly \cite{elhoseiny2013write,romera2015embarrassingly} or non-linearly via deep neural networks~\cite{lei2015predicting} or kernels~\cite{Elhoseiny_Write_2016}). 
At test-time, classification is performed by associating the image to the class with the highest similarity to the corresponding class-level text. 
%
Recently, Reed \etal~\cite{ReedASL16} showed that by collecting 10 sentences per-image, their sentence-based approach can outperform  attribute-based alternatives on CUB. 


In contrast to these approaches, we directly map external domain knowledge (text-based or otherwise) to internal components (neurons) of deep neural networks rather than learning associative mappings between images and text -- providing interpretability for our novel classifiers. \\[-20pt]

\section{Neuron Importance-Aware Weight Transfer (NIWT)}

\begin{figure}[t]
\centering
\begin{subfigure}{0.315\textwidth}
\centering
\includegraphics[height=1in]{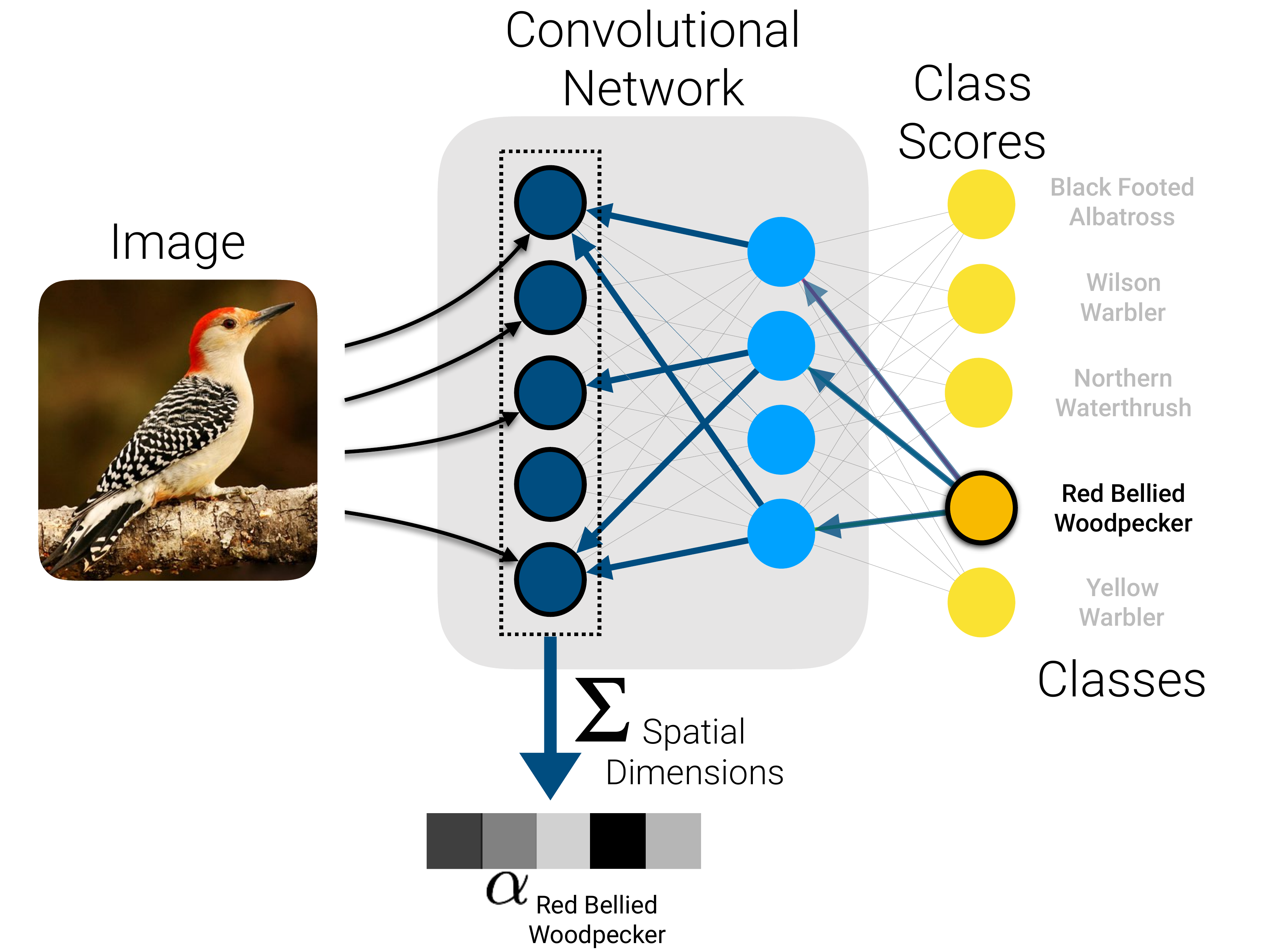}\\
\caption{}
\label{fig:app1}
\end{subfigure}
\begin{subfigure}{0.315\textwidth}
\centering
\includegraphics[height=1in]{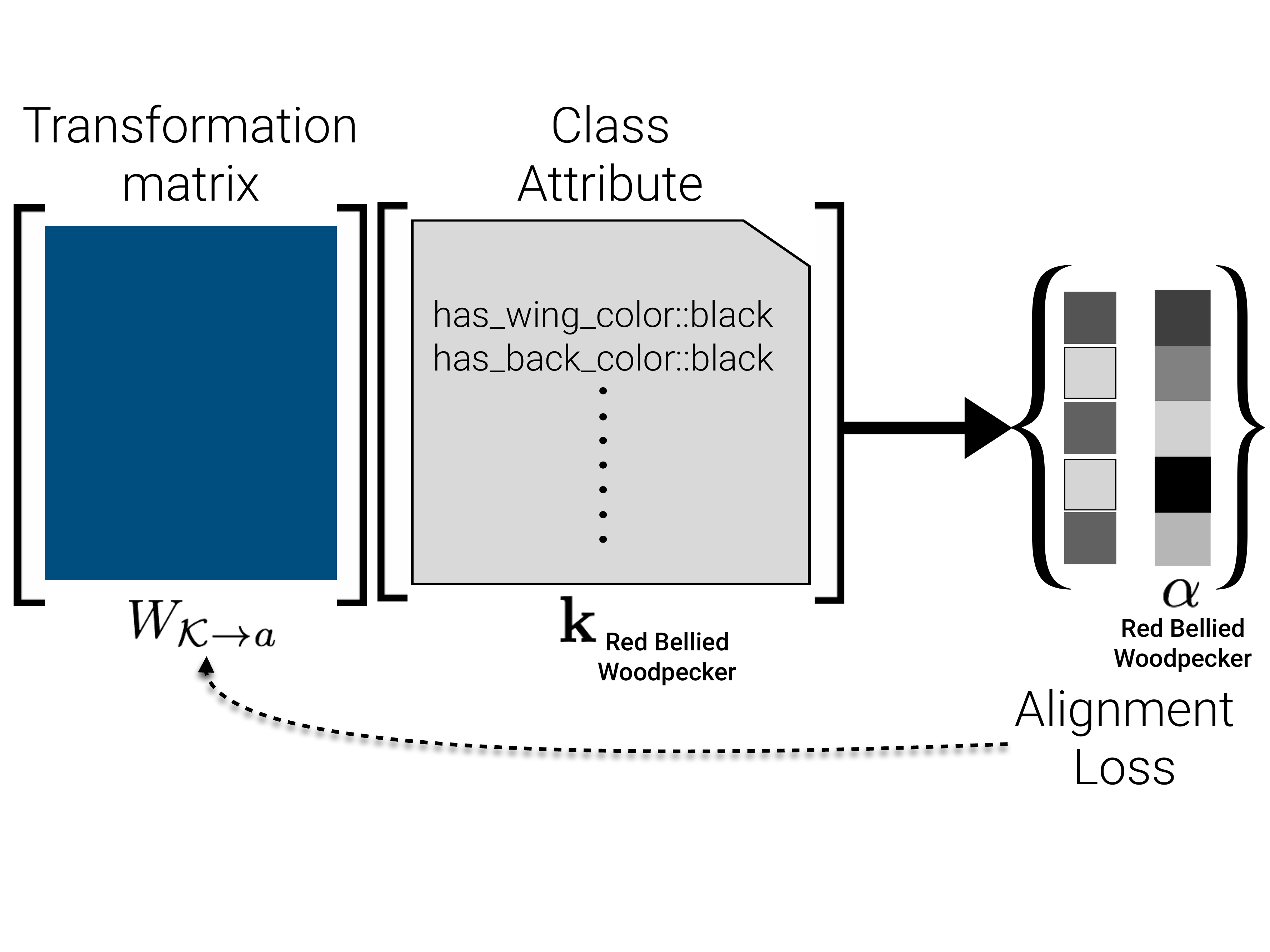}\\
\caption{}
\label{fig:app2}
\end{subfigure}
\begin{subfigure}{0.315\textwidth}
\centering
\includegraphics[height=1in]{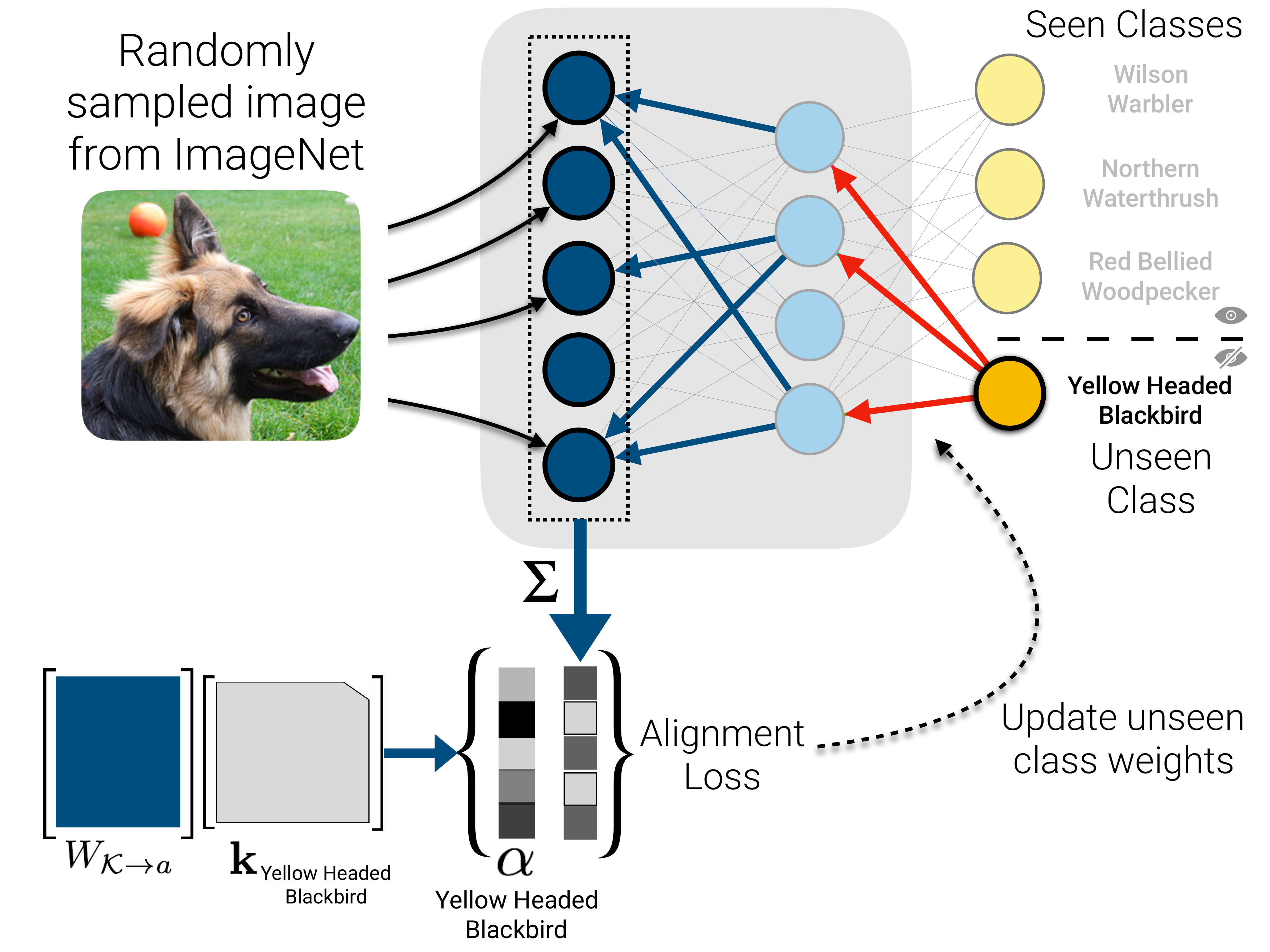}\\
\caption{}
\label{fig:app3}
\end{subfigure}\\[-8pt]
\caption{Our Neuron Importance-Aware Weight Transfer (NIWT) approach can be broken down in to three stages. a) class-specific neuron importances are extracted for seen classes at a fixed layer, b) a linear transform is learned to project free-form domain knowledge to these extracted importances, and c) weights for new classifiers are optimized such that neuron importances match those  predicted by this mapping for unseen classes. \\[-30pt]}
\end{figure}

In this section, we describe our Neuron Importance-Aware Weight Transfer (NIWT) approach to zero-shot learning. At a high level, NIWT maps free-form domain knowledge to neurons within a deep network and then learns classifiers based on novel class descriptions which respect these groundings. Concretely, NIWT consists of three steps: 
(1) estimating the importance of individual neuron(s) at a fixed layer w.r.t. the decisions made by the network for the seen classes (see Figure \ref{fig:app1}),
(2) learning a mapping between domain knowledge and these neuron-importances (see Figure \ref{fig:app2}), and (3) optimizing classifier weights with respect to predicted neuron-importances for unseen classes (see Figure \ref{fig:app3}).
We discuss each stage in the following sections.
\\[-20pt]

\subsection{Preliminaries: Generalized Zero-Shot Learning (GZSL)}
\label{sec:gzsl}

Consider a dataset ${\cal D} = \{(x_i, y_i)\}_{i=1}^N$  comprised of example input-output pairs from a set of  \emph{seen classes} ~${\cal S}=\{1,\dots,s\}$ and \emph{unseen classes} ~${\cal U}=\{s{+}1,\dots,s{+}u\}$. For convenience, we use the subscripts $\cal S$ and $\cal U$ to indicate subsets corresponding to seen and unseen classes respectively, \eg ${\cal D}_S = \{ (x_i, y_i) ~|~ y_i \in {\cal S}\}$. Further, assume there exists domain knowledge ${\cal K} = \{k_1, ...,  k_{s+u}\}$  corresponding to each class 
(\eg class level attributes or natural language descriptions). 
Concisely, the goal of generalized zero-shot learning is then to learn a mapping $f: {\cal X} \rightarrow {\cal S} \cup {\cal U}$ from the input space ${\cal X}$ to the combined set of seen and unseen class labels using only the domain knowledge ${\cal K}$ and instances ${\cal D}_{\cal S}$ belonging to the seen classes. \\[-25pt]

\subsection{Class-dependent Neuron Importance}
\label{sec:importance}

Class descriptions capture salient concepts about the content of corresponding images -- for example, describing the coloration and shape of a bird's head. 
Similarly, a classifier must also learn discriminative visual concepts in order to succeed; however, these concepts are not grounded in human interpretable language. In this stage, we identify neurons corresponding to these discriminative concepts before aligning them with domain knowledge in \secref{sec:map_knowledge}.

Consider a deep neural network $\mathtt{NET}_\mathcal{S}( \cdot )$ trained for classification which predicts scores $\{o_c ~|~ c \in {\cal S}\}$ for seen classes $\cal{S}$. One intuitive measure of a neuron $n$'s importance to the final score $o_c$ is simply the gradient of $o_c$ with respect to the neuron's activation $a^n$ (where $n$ indexes the channel dimension).
For networks containing convolutional units (which are replicated spatially), we follow \cite{gradcam_arxiv} and simply compute importance as the mean gradient (along spatial dimensions), writing the neuron importance $\alpha^n_c$ as \\[-10pt]
\begin{equation}
    \alpha{}_{c}^n =
    \overbrace{
        \frac{1}{HW}\sum_{i=1}^{H}\sum_{j=1}^{W}
    }^{\text{global average pooling}} \mkern-65mu
    \underbrace{
        \vphantom{\sum_{i}\sum_{j}} \frac{\partial o_c}{\partial a_{ij}^{n}}
    }_{\text{gradients via backprop}}
\end{equation}
where $a^n_{i,j}$ is the activation of neuron $n$ at spatial position $i,j$. For a given input, the importance of every neuron in the network can be computed for a given class via a single backward pass followed by a global average pooling operation for convolutional units. In practice, we focus on $\alpha$'s from single layers in the network in our experiments. We note that other measures of neuron importance have been proposed \cite{nisp_prune,apple_konam} in various contexts; however, this simple gradient-based importance measure has some notable properties which we leverage. 

 Firstly, we find gradient-based importance scores to be quite consistent across images of the same class despite the visual variation between instances, and likewise to correlate poorly across classes. 
%
%
To assess this quantitatively, we computed $\alpha$'s for neurons in the final convolutional layer of a convolutional neural network trained on a fine-grained multi-class task (\texttt{conv5-3} of VGG-16 \cite{simonyan2014very} trained on AWA2 \cite{Xian_2017_CVPR}) for 10,000 randomly selected images. We observed an average rank correlation of 0.817 
for instances within the same class 
and 0.076 across pairs of classes. 
This relative invariance of $\alpha$'s to intra-class input variation may be due in part to the piece-wise linear decision boundaries in networks using ReLU \cite{nair2010rectified} activations. As shown in \cite{novak2018sensitivity}, transitions between these linear regions are much less frequent between same-class inputs than across classes. Within the same linear region, activation gradients (and hence $\alpha$'s) are trivially identical.

Secondly, this measure is fully differentiable with respect to model parameters which we use to learn novel classifiers with gradient methods (see \secref{sec:map_weight}) \\[-20pt].

\subsection{Mapping Domain Knowledge to Neurons} 
\label{sec:map_knowledge}
Without loss of generality, consider a single layer $L$ within $\mathtt{NET}_\mathcal{S}( \cdot )$. Given an instance $(x_i, y_i) \in {\cal D}_{\cal S}$, let $\mathbf{a}_c = \{\alpha^n_{c} ~|~ n \in L \}$ be a vector of importances computed for neurons in $L$ with respect to class $c$ when $x_i$ is passed through the network.  In this section, we learn a simple linear mapping from domain knowledge to these importance vectors -- aligning interpretable semantics with individual neurons.

We first compute the importance vector $\mathbf{a}_{y_i}$ for each seen class instance $(x_i,y_i)$ and match it with the domain knowledge representation $k_{y_i}$ of the corresponding class. 
Given this dataset of $(\mathbf{a}_{y_i}, k_{y_i})$ pairs, we
learn a linear transform $W_{ {\cal K} \rightarrow a}$ to map domain knowledge to importances.
As importances are gradient based, we penalize errors in the predicted importances based on cosine distance -- emphasizing alignment over magnitude. We minimize the cosine distance loss as
%
%
\begin{equation}
{\cal L}(\mathbf{a}_{y_i}, \mathbf{k}_{y_i}) =  1 - \frac{\left(W_{ {\cal K} \rightarrow a} \cdot \mathbf{k}_{y_i}\right)\cdot \mathbf{a}_{y_i}}{\lVert W_{ {\cal K} \rightarrow a} \cdot \mathbf{k}_{y_i}\rVert~\lVert \mathbf{a}_{y_i}\rVert}, 
\end{equation}
via gradient descent to estimate $W_{ {\cal K} \rightarrow a}$.  We stop training when average rank-correlation of predicted and true importance vectors stabilizes for a set of held out validation classes from $\cal S$. 

Notably, this is a many-to-one mapping with the domain knowledge of one class needing to predict many different importance vectors. Despite this, this mapping achieves average rank correlations of 0.2 to 0.5 for validation class instances. We explore the impact of error in importance vector prediction on weight optimization in \secref{sec:map_weight}. 
We also note that this simple linear mapping can also be learned in an inverse fashion,
mapping neuron importances back to semantic concepts within the domain knowledge (which we explore in \secref{sec:explanations}) \\[-20pt].

\subsection{Neuron Importance to Classifier Weights} 
\label{sec:map_weight}
In this section, we use predicted importances to learn classifiers for the unseen classes. As these new classifiers will be built atop the trained seen-class network $\mathtt{NET}_\mathcal{S}$, we modify $\mathtt{NET}_\mathcal{S}$ to extend the output space to include the unseen class -- expanding the final fully-connected layer to include additional neurons with weight vectors $\mathbf{w}^1, \dots, \mathbf{w}^u$ for the unseen classes such that the network now additionally outputs scores $\{o_c ~|~ c \in {\cal U}\}$. We refer to this expanded network as $\mathtt{NET}_{\mathcal{S}\cup\mathcal{U}}$. At this stage, the weights for the unseen classes are sampled randomly from a multivariate normal distribution with parameters estimated from the seen class weights and as such the output scores are uncalibrated and uninformative.

Given the learned mapping $W_{\mathcal K \rightarrow A}$ and unseen class domain knowledge $\mathcal{K}_\mathcal{U}$, we can predict unseen class importances $A_\mathcal{U} = \{\mathbf{a}_1,...,\mathbf{a}_u\}$ with the importance vector for unseen class $c$ predicted as $\mathbf{a}_c = W_{\mathcal K \rightarrow a}\mathbf{k}_c$. For a given input, we can compute importance vectors $\hat{a}_c$ for each unseen class $c$. As $\hat{\mathbf{a}}^c$ is a function of the weight parameters $\mathbf{w}_c$, we can simply supervise $\hat{\mathbf{a}}_c$ with the predicted importances $\mathbf{a}_c$ and optimize $w^c$ with gradient descent -- minimizing the cosine distance loss between predicted and observed importance vectors. However, the cosine distance loss does not account for scale and without regularization the scale of weights (and as consequence the outputs) of seen and unseen classes might vary drastically, resulting in bias towards one set or the other. 

To address this problem, we introduce a $L_2$ regularization term which constrains the learned unseen weights to be a similar scale as the mean of seen weights $\overline{\mathbf{w}}_\mathcal{S}$. We write the final objective as 
\begin{equation}
{\cal L}(\hat{\mathbf{a}}_c, \mathbf{a}_c) =  1 - \frac{\hat{\mathbf{a}}_c\cdot \mathbf{a}_c}{\lVert \hat{\mathbf{a}}_c\rVert~\lVert \mathbf{a}_c\rVert} + \lambda\lVert\mathbf{w}_c - \overline{\mathbf{w}}_\mathcal{S}\rVert, 
\end{equation}
where $\lambda$ is controls the strength of this regularization. We examine the effect of this trade-off in \secref{sec:reg}, finding training to be robust to a wide range of $\lambda$ values.
We note that as observed importances $\mathbf{a}^c$ are themselves computed from network gradients, updating weights based on this loss requires computing a Hessian-vector product; however, this is relatively efficient as the number of weights for each unseen class is small and independent of those for other classes. 

\vspace{5pt}
\par \noindent
\textbf{Training Images.} Note that to perform the optimization described above, we need to pass images through the network to compute importance vectors. We observe importances to be only weakly correlated with image features and find they can be computed for any of the unseen classes irrespective of the input image class -- as such, we find simply inputing images with natural statistics to be sufficient. 
Specifically, we pair random images from ImageNet~\cite{imagenet_cvpr09} with random tuples $(\hat{\mathbf{a}}_c, \mathbf{k}_c)$ to perform the importance to weight optimization. \\[-20pt]

\section{Experiments}

In this section, we evaluate our approach on generalized zero-shot learning (GZSL) (\secref{sec:gzsl_exp}) and present analysis for each stage of NIWT (\secref{sec:analysis}). \\[-20pt]

\subsection{Experimental Setting}
\label{sec:gzsl_exp}
\par \noindent
\textbf{Datasets and Metrics.}
We conduct our GZSL experiments on the\\[-10pt]
\begin{compactitem}
\item \textbf{Animals with Attributes 2 (AWA2) \cite{Xian_2017_CVPR}} -- The AWA2 dataset consists of 37,322 images of 50 animal species (on average 764 per class but with a wide range). Each class is labeled with 85 binary and continuous attributes.\\[-8pt]
\item \textbf{Caltech-UCSD Birds 200 (CUB) \cite{WahCUB_200_2011}} -- The CUB dataset consists of 11788 images corresponding to 200 species of birds. Each image and each species has been annotated with 312 binary and continuous attribute labels respectively. These attributes describe fine-grained physical bird features such as the color and shape of specific body parts. Additionally, each image is associated with 10 human captions \cite{ReedASL16}.\\[-8pt]
\end{compactitem}

\par\noindent For both datasets, we use the GZSL splits proposed in \cite{Xian_2017_CVPR} which ensure  that no unseen class occurs within the  ImageNet \cite{imagenet_cvpr09} dataset which is commonly used for training classification networks for feature extraction. As in \cite{xian2016latent}, we evaluate our approach using class-normalized accuracy computed over both seen and unseen classes (\ie 200-way for CUB) -- breaking the results down into unseen accuracy $\mathtt{Acc}_\mathcal{U}$, seen accuracy $\mathtt{Acc}_\mathcal{S}$, and the harmonic mean between them $\mathtt{H}$.

\vspace{5pt}
\par \noindent
\textbf{Models.} We experiment with ResNet101 \cite{he_cvpr15} and VGG16 \cite{simonyan_arxiv14} models pretrained on ImageNet \cite{imagenet_cvpr09} and fine-tuned on the seen classes. For each, we train a version by finetuning all layers and another by updating only the final classification weights.
Compared to ResNet, where we see sharp declines for fixed models (60.6\% finetuned vs 28.26\% fixed for CUB and 90.10\% vs 70.7\% for AWA2), VGG achieves similar accuracies for both finetuned and fixed settings (74.84\% finetuned vs 66.8\% fixed for CUB and 92.32\% vs 91.44\% for AWA2).
We provide more training details in the Appendix. 

\vspace{5pt}
\par \noindent
\textbf{NIWT Settings.} To train the domain knowledge to importance mapping we hold out five seen classes and stop optimization when rank correlation between observed and predicted importances is highest. For attribute vectors, we use the class level attributes directly and for captions on CUB we use average word2vec embeddings\cite{Mikolov_2013} for each class.
When optimizing for weights given importances, we stop when the loss fails to improve by 1\% over 40 iterations. We choose values of $\lambda$ (between $1e^{-5}$ to $1e^{-2}$), learning rate ( $1e^{-5}$ to $1e^{-2}$) and the batch size ($\{ 16, 32, 64 \}$) by  grid search on $\mathtt{H}$ for a disjoint set of validation classes sampled from the seen classes of the proposed splits~\cite{Xian_2017_CVPR} based  (see Table.~\ref{tab:gzsl}).

\vspace{5pt}
\par \noindent
\textbf{Baselines.}
We compare NIWT with a number of well-performing zero-shot learning approaches based on learning joint embeddings of image features and class information. 
Methods like ALE \cite{akata2016label}
focus on learning compatibility functions for class labels and visual features using some form of ranking loss. In addition to comparing with ALE as reported in \cite{Xian_2017_CVPR}, we also compare with settings where the hyper-parameters have been directly tuned on the test-set.

We also compare against the recent Deep Embedding approach of \cite{zhang2017learning} which also leverages deep networks, jointly aligning domain knowledge with deep features end-to-end. 
For both of the mentioned baselines, we utilize code provided by the authors and report results by directly tuning hyper-parameters on the test-set so as to convey an upper-bound of performance.
\begin{table}[t] \footnotesize
\renewcommand*{\arraystretch}{1.18}
\setlength{\tabcolsep}{6pt}
\begin{center}
\resizebox{\columnwidth}{!}{
\begin{tabular}{l l l  c  c c c  c  c c c}
\toprule 
& & & & \multicolumn{3}{c}{AWA2 \cite{Xian_2017_CVPR}} & & \multicolumn{3}{c}{CUB \cite{WahCUB_200_2011}}\\
& & Method & & $\mathtt{Acc}_{\cal{U}}$ & $\mathtt{Acc}_{\cal{S}}$ & $\mathtt{H}$ & & $\mathtt{Acc}_{\cal{U}}$ & $\mathtt{Acc}_{\cal{S}}$ & $\mathtt{H}$\\
\midrule
\multirow{8}{*}{\rotatebox{90}{\centering ResNet101 \cite{he_cvpr15}}} & 
\multirow{4}{*}{\rotatebox{90}{\centering Fixed}} 
& ALE \cite{akata2016label}$^1$ && 14.0 & {81.8} & 23.9 & & 23.7 & \textbf{62.8} & \textbf{34.4}\\
& & ALE \cite{akata2016label}$^2$ && 20.9 & \textbf{88.8} & 33.8 & & \textbf{24.7} & 62.3 & \textbf{34.4}\\
& & Deep Embed. \cite{zhang2017learning}$^2$ && \textbf{28.5}	& 82.3 & \textbf{42.3} && 22.3 & 45.1 & 29.9\\
& & NIWT-Attributes && 21.6 &	37.8 &	27.5 && 10.2 & 57.7 & 17.3\\
\cline{3-11}
& \multirow{4}{*}{\rotatebox{90}{FT}} 
& ALE \cite{akata2016label}$^2$ && 22.7 & \textbf{75.1} & 34.9 & & 24.1 & \textbf{60.8} & \textbf{34.5}\\
& & Deep Embed. \cite{zhang2017learning}$^2$ && 21.5 & {59.6} & 31.6 && \textbf{24.7} & {57.4} & \textbf{34.5} \\
& & NIWT-Attributes && \textbf{42.3} & 38.8 & \textbf{40.5} && {20.7} & {41.8} & 27.7\\
\cline{3-11}
& & NIWT-Caption &&&  N/A &&& 22.1 & 25.7 & 23.8\\
\bottomrule\\[-10pt]
\multirow{7}{*}{\rotatebox{90}{\centering VGG16 \cite{simonyan_arxiv14}}} & 
\multirow{4}{*}{\rotatebox{90}{\centering Fixed}} 
& ALE \cite{akata2016label}$^2$ && 17.9 & \textbf{84.3} & 29.5 && 22.2 & \textbf{54.8} & \textbf{31.6}\\
& & Deep Embed. \cite{zhang2017learning}$^2$ && \textbf{28.8} & 81.7 & \textbf{42.6} && \textbf{24.1} & 45.2 & 31.5 \\
& & NIWT-Attributes && 43.8 & 30.7	& 36.1 && 17.0 & 54.6 & 26.7\\
\cline{3-11}
& \multirow{3}{*}{\rotatebox{90}{\centering FT}}
& ALE \cite{akata2016label}$^2$ && 16.9 & \textbf{91.5} & 28.5&& 25.3 & \textbf{62.6} & {36.0}\\
& & Deep Embed. \cite{zhang2017learning}$^2$  && 26.6 & {83.3} & 38.2 && 27.0 & {49.7} & 35.0\\
& & NIWT-Attributes  && \textbf{35.3} & {75.5} & \textbf{48.1} && \textbf{31.5} & 44.9 & \textbf{37.0}\\
\cline{3-11}
& & NIWT-Caption  && &  N/A &&& 15.9 &	46.5 &	23.6\\
\bottomrule
\end{tabular}}\\[5pt]
\caption{Generalized Zero-Shot Learning performances on the proposed splits \cite{Xian_2017_CVPR} for AWA2 and CUB. We report class-normalized accuracies on seen and unseen classes and harmonic mean.
$^1$ reproduced from \cite{Xian_2017_CVPR}. 
$^2$ based on code provided by the authors by tuning hyper-parameters on the test-set to convey an upper-bound of performance. \\[-40pt]}
\label{tab:gzsl}
\end{center}
\end{table}

\subsection{Results} 
We show results in Table \ref{tab:gzsl} for AWA2 and CUB using all model settings. There are a number of interesting trends to observe:
\begin{compactenum}[1.]
\item \textbf{NIWT sets the state of the art in generalized zero-shot learning.} For both datasets, NIWT-Attributes based on VGG establishes a new state of the art for harmonic mean (48.1\% for AWA2 and 37.0\% for CUB). For AWA2, this corresponds to a $\sim10\%$ improvement over prior state-of-the-art which is based on deep feature embeddings. These results imply that mapping domain knowledge to internal neurons can lead to improved results. 
\\[-8pt]
\item \textbf{Seen-class finetuning yields improved harmonic mean $\mathtt{H}$.} 
%
%
For CUB and AWA2, finetuning the VGG network on seen class images offers significant gains for NIWT (26.7\%$\rightarrow$37.0\% $\mathtt{H}$ and 36.1\%$\rightarrow$48.1\% $\mathtt{H}$ respectively); finetuning ResNet sees similar gains (17.3\%$\rightarrow$27.7\% $\mathtt{H}$ on CUB and 27.5\%$\rightarrow$40.5 \%$\mathtt{H}$ on AWA2). Notably, these trends seem inconsistent for the compared methods.\\[-8pt]
\item \textbf{NIWT effectively grounds both attributes and free-form language.}  We see strong performance both for attributes and captions across both networks (37.0\% and 23.6\% $\mathtt{H}$ for VGG and 27.7\% and 23.8\% $\mathtt{H}$ for ResNet). We note that we use relatively simple, class-averaged representations for captioning which may contribute to the lower absolute performance.\\[-22pt] 
\end{compactenum}


\section{Analysis}
\label{sec:analysis}

To better understand the different stages of NIWT, we perform a series of experiments to analyze and isolate individual components in our approach. \\[-20pt]

\subsection{Effect of Regularization Coefficient $\lambda$.}
\label{sec:reg}

\begin{figure}[t]
\centering
\begin{subfigure}{0.45\textwidth}
\centering
\includegraphics[height=1.6in]{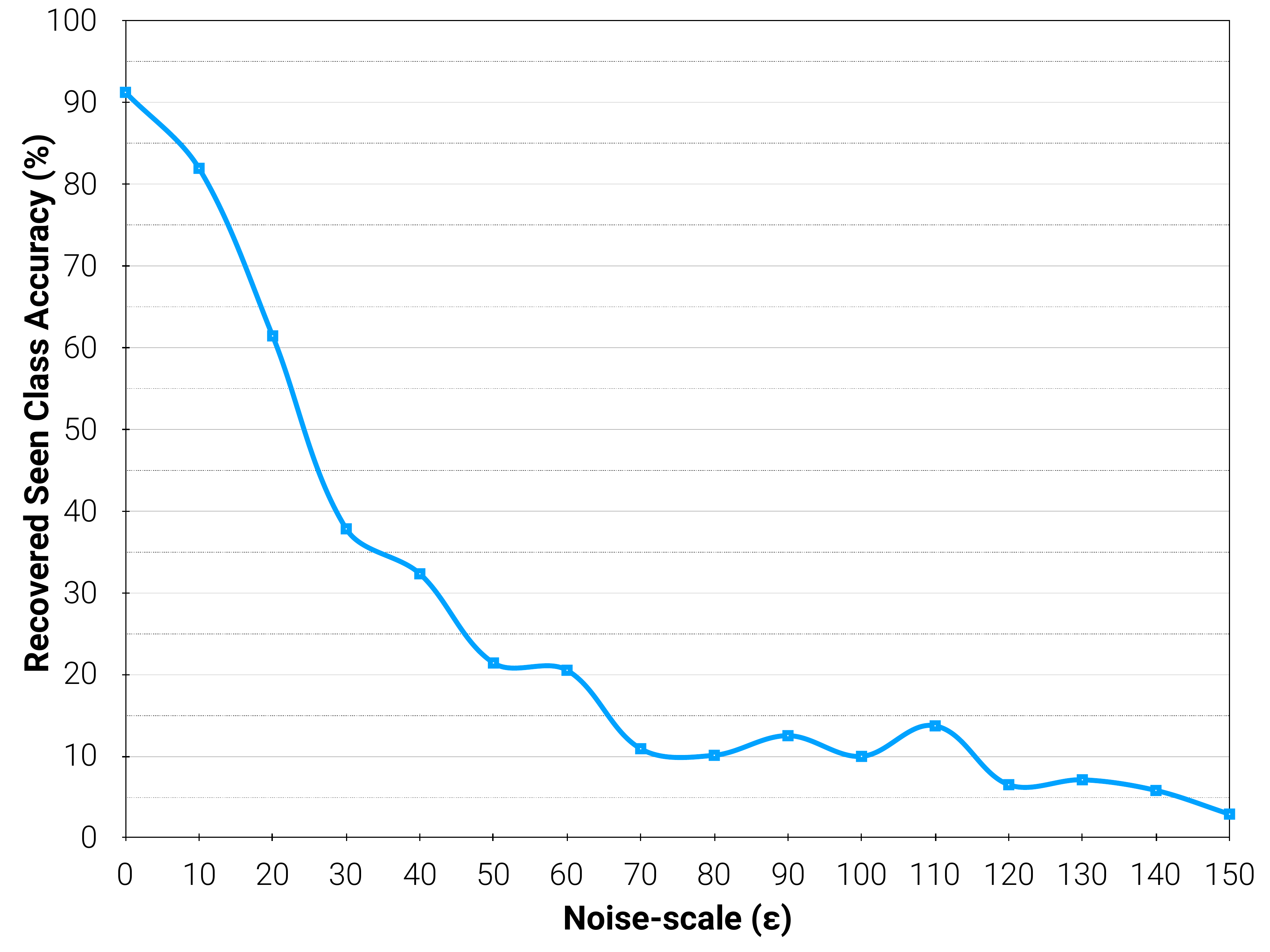}\\
\caption{Noise Tolerance ($\epsilon$)}
\label{fig:noise_recovery}
\end{subfigure}\hfill
\begin{subfigure}{0.45\textwidth}
\centering
\includegraphics[clip=true, trim=25px 5px 10px 0px, height=1.6in]{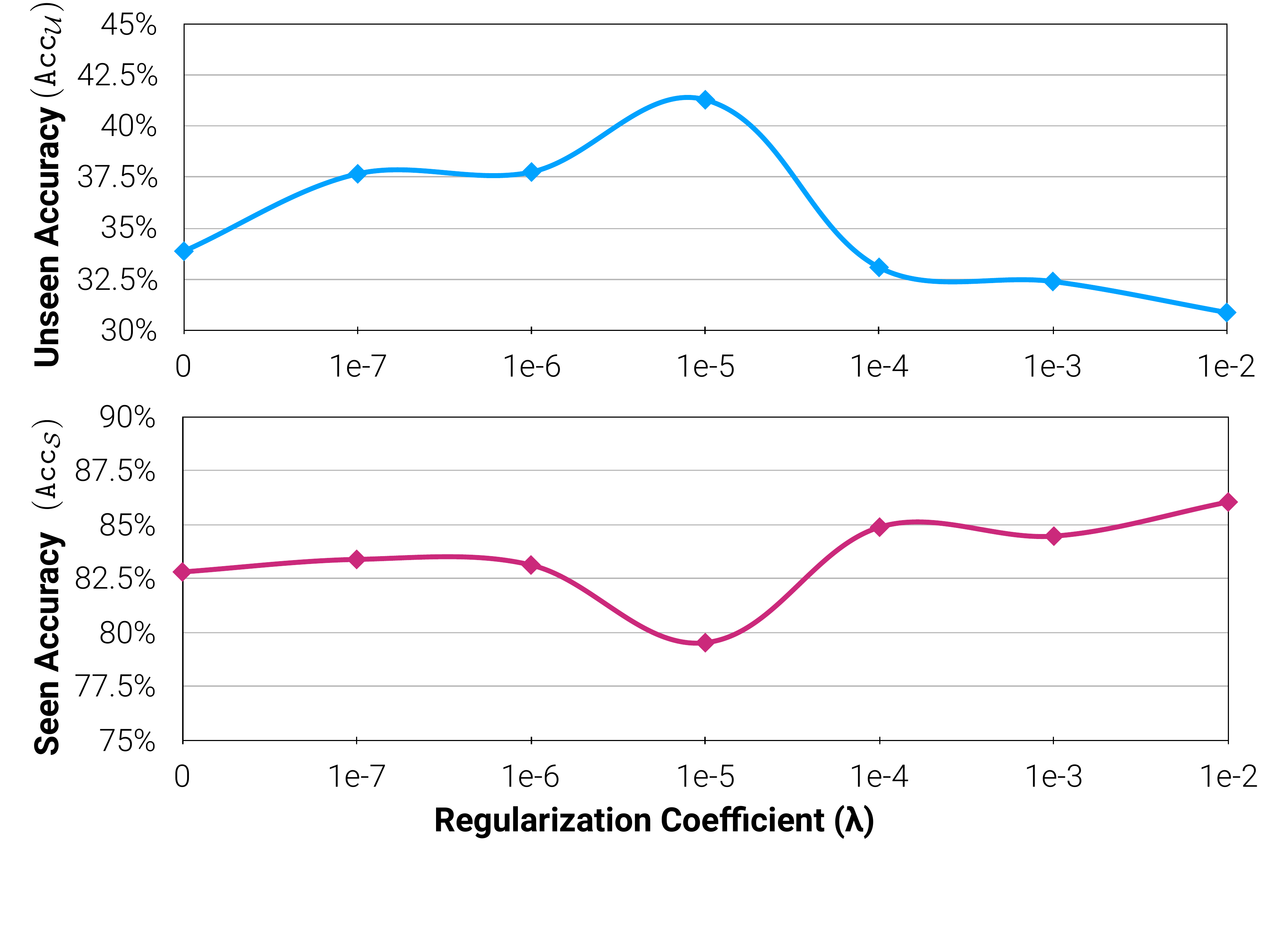}\\
\caption{Regularizer Sensitivity ($\lambda$)}
\label{fig:reg}
\end{subfigure}
\caption{Analysis of the importance vector to weight optimization for VGG-16 trained on AWA2 (a). We find that ground-truth weights can be recovered for a pre-trained network even in the face of high magnitude noise. (b) We also show the importance of the regularization term to final model performance. \\[-60pt]}
\end{figure}

One key component to our importance to weight optimization is the regularizer which enforces that learned unseen weights be close to the mean seen weight -- avoiding arbitrary scaling of the learned weights and the bias this could introduce. To explore the effect of the regularizer, we vary the coefficient $\lambda$ from 0 to $1e^{-2}$. Figure \ref{fig:reg} shows the final seen and unseen class-normalized accuracy for the AWA2 dataset at convergence for different $\lambda$'s. 

Without regularization ($\lambda{=}0$) the unseen weights tend to be a bit too small and achieve an unseen accuracy of only 33.9\% on AWA2. As $\lambda$ is increased the unseen accuracy grows until peaking at $\lambda{=}1e^{-5}$ with an unseen accuracy of 41.3\% -- an improvement of over 8\% from the unregularized version! Of course, this improvement comes with a trade-off in seen accuracy of about 3\% over the same interval. 
As $\lambda$ grows larger ${>}1e^{-4}$, the regularization constraint becomes too strong and NIWT has trouble learning anything for the unseen classes. \\[-20pt]

\subsection{Noise Tolerance in Neuron Importance to weight optimization}

One important component of NIWT is the ability to ground concepts learnt by a convolutional network in some referable domain. Due to the inherent noise involved in this mapping $\mathit{W}_{\mathcal{K} \rightarrow \mathit{A}}$, the classifiers obtained for unseen classes in the expanded network $\mathtt{NET}_{\mathcal{S} \cup \mathcal{U}}$ are not perfect. In order to judge the capacity of the optimization procedure, we experiment with a toy setting where we initialize an unseen classifier head with the same dimensionality as the seen classes and try to explicitly recover the seen class weights with supervision only from the \textit{oracle} $\mathbf{a}_c$ obtained from the seen classifier head for the seen classes. To simulate for the error involved in estimating $\mathbf{a}_c$, we add increasing levels of zero-centered gaussian noise and study recovery performance in terms of accuracy of the recovered classifier head on the seen-test split. That is, the supervision from importance vectors is constructed as follows:\\[-10pt]
\begin{equation}
\tilde{\mathbf{a}}_c = \mathbf{a}_c + \epsilon \overline{||\mathbf{a}_c||}_{1} \mathcal{N}(0,I) 
\end{equation}
%
We operate at different values of $\epsilon$, characterizing different levels of corruption of the supervision from $\mathbf{a}_c$ and observe recovery performance in terms of accuracy of the recovered classifier head. \ref{fig:noise_recovery} shows the effect of noise on the ability to recover seen classifier weights (\texttt{fc7}) for a VGG-16 network trained on 40 seen classes of AWA2 dataset with the same objective as the one used for unseen classes.


In the absence of noise over $\mathbf{a}_c$ supervision, we find that we are exactly able to recover the seen class weights and are able to preserve the pre-trained accuracy on seen classes. 
Even with a noise-level of $\epsilon{=}10$ (or adding noise with a magnitude 10x the average norm of $\mathbf{a}_c$), we 
observe only minor reduction in the accuracy of the recovered seen class weights.
As expected, this downward trend continues as we increase the noise-level until we reach almost chance-level performance on the recovered classifier head. This experiment shows that the importance vector to weights optimization is quite robust even to fairly extreme noise. \\[-20pt]


\subsection{Network Depth of Importance Extraction.}

In this section, we explore the sensitivity of NIWT with respect to the layer from which we extract importance vectors in the convolutional network. As an experiment (in addition to Table~\ref{tab:gzsl}) we evaluate NIWT on AWA2 with importance vectors extracted at different convolutional layers of VGG-16. We observe that out of those we experimented with \texttt{conv5\_3} performs the best with $\mathtt{H} = 48.1$ followed by \texttt{conv4\_3} ($\mathtt{H} = 39.3$), \texttt{conv3\_3} ($\mathtt{H} = 35.5$), \texttt{conv2\_2} ($\mathtt{H} = 23.8$) and \texttt{conv2\_2} ($\mathtt{H} = 20.8$). We also experimented with the fully-connected layers \texttt{fc6} and \texttt{fc7} resulting in values of $\mathtt{H}$ being $40.2 $ and $1$ respectively. 

Note that performing NIWT on importance vectors extracted from the penultimate layer \texttt{fc7} is equivalent to learning the unseen head classifier weights directly from the domain space representation ($\mathbf{k}_c$).
 Consistent with our hypothesis, this performs very poorly across all the metrics with almost no learning involved for the unseen classes at all. Though we note that this may be due to the restricted capacity of the linear transformation $\mathit{W}_{\mathcal{K}\rightarrow\mathit{A}}$ involved in the process. \\[-20pt]




\subsection{Importance to Weight Input Images}
\begin{minipage}{0.55\textwidth}
We evaluate performance with differing input images during weight optimization (random noise images, ImageNet images, and seen class images). We show results of each in Table \ref{tab:data}.
As expected, performance improves as input images more closely resemble the unseen classes; however, we note that learning occurs even with random noise images.
\end{minipage}\hfill
\resizebox{0.4\textwidth}{!}{\begin{minipage}{0.4\textwidth}
\setlength{\tabcolsep}{1pt}
        \begin{tabular}{@{} l  c  c  c  c  c@{}}
		\toprule
		Sampling Mode & $\mathtt{Acc}_{\mathcal{U}}$ & $\mathtt{Acc}_{\mathcal{S}}$ & $\mathtt{H}$ \\
		\midrule
		Random Normal & 23.9 & 41.0 & 30.2 \\
		ImageNet & 31.5 & 44.9 & 37.0 \\
		Seen-Classes & 36.4 & 40.0 & 38.1 \\
		\bottomrule
		\end{tabular}
      \captionof{table}{{Results by sampling images from different sets for NIWT-Attributes on VGG-CUB.}}
      \label{tab:data}
\end{minipage}}

\section{Explaining NIWT \\[-100pt]}
\label{sec:explanations}
In this section we demonstrate how we can use NIWT to provide visual and textual explanations for the decisions made by the newly learned classifiers on the unseen classes. In addition to the visual explanations provided by Grad-CAM~\cite{gradcam_arxiv}, we utilize a mapping (similar to the one in Sec.~\ref{sec:map_knowledge}) learned in the inverse direction -- $\mathit{W}_{a \rightarrow \mathcal{K}}$, i.e., neuron-importance(s) $\mathbf{a}_c$ to domain knowledge $\mathcal{K}$ to ground the predictions made in the textual domain used as external knowledge. Since this mapping explicitly grounds the \emph{important} neurons in the interpretable domain, we automatically obtain neuron names. 



\noindent
\textbf{Visual Explanations.}
Since NIWT learns the classifier associated with the unseen classes as an extension to the existing deep network for the seen classes, it preserves the end-to-end differentiable pipeline for the novel classes as well. This allows us to directly use any of the existing deep network interpretability mechanisms to visually explain the decisions made at inference. 
We use Grad-CAM~\cite{gradcam_arxiv} on instances of unseen classes to visualize the support for decisions (see Fig.~\ref{fig:explanations}) 
made by the network with NIWT learnt classification weights.



\noindent
\textbf{Evaluating Visual Explanations.}
Quantitatively, we evaluate the generated maps for both seen and unseen classes by the mean fraction of the Grad-CAM activation present inside the bounding box annotations associated with the present objects. On seen classes, we found this number to be $0.80\pm 0.008$ versus $0.79\pm 0.005$ for the unseen classes on  CUB -- indicating that the unseen classifier learnt via NIWT is indeed capable of focusing on relevant regions in the input image while making a prediction.

\par \noindent
\textbf{Textual Explanations.}
In Sec.~\ref{sec:map_knowledge}, we learned a mapping $W_{{\cal K} \rightarrow a}$ to embed the external domain knowledge (attributes or captions) into the neurons of a specific layer of the network. Similarly, by learning an inverse mapping from the neuron importances to the attributes (or captions), we can ground the former associated with a prediction in a human-interpretable domain. We utilize such a inverse mapping to obtain scores in the attribute-space (given $\mathbf{a}_c$) and retrieve the top-k attributes as explanations.
A high scoring $\mathbf{k}_c$ retrieved via $W_{ a \rightarrow {\cal K}}$ from a certain $\mathbf{a}_c$ emphasizes the relevance of that attribute for the corresponding class $c$. This helps us ground the class-score decisions made by the learnt unseen classifier head in the attribute space, thus, providing an explanation for the decision. \\[-20pt]


\par \noindent
\textbf{Evaluating Textual Explanations.} We evaluate the fidelity of such 
textual explanations by the percentage of associated ground truth attributes captured in the top-k generated explanations on a per instance level
-- $83.9\%$ on CUB using a VGG-16 network. 
Qualitative results in \figref{fig:explanations} show visual and textual explanation(s) demonstrating the discriminative attributes learned by the model for predicting the given target category. \\[-20pt]

\begin{minipage}{0.94\textwidth}
    \begin{center}
  \centering
  \includegraphics[width=\textwidth]{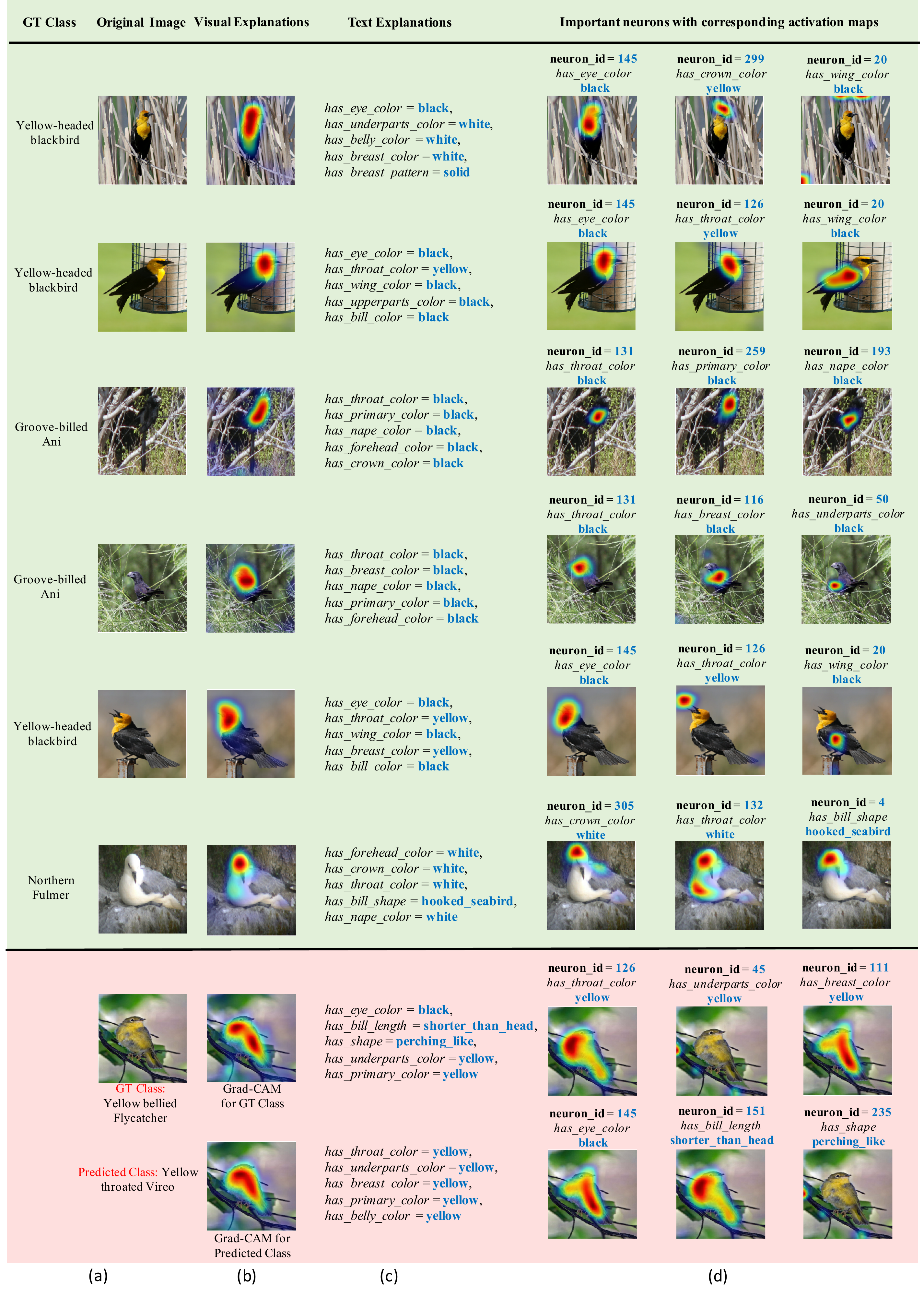}\\
	\captionof{figure}{Success and failure cases for unseen classes using explanations for NIWT: Success cases: (a) the ground truth class and image, (b) Grad-CAM visual explanations for the GT category, (c) textual explanations obtained using the inverse mapping from $\mathbf{a}_c$ to domain knowledge, (d) most important neurons for this decision, their names and activation maps. The last 2 rows show failure cases, where the model predicted a wrong category.  We show Grad-CAM maps and textual explanations for both the ground truth and predicted category. By looking at the explanations for the failure cases we can see that the model's mistakes are not completely unreasonable. \\[-20pt]}
	\label{fig:explanations}
\end{center}
\end{minipage}


\par \noindent
\textbf{Neuron Names and Focus.}
Neuron names are referable groundings of concepts captured by a deep convolutional network. Unlike previous approaches, we obtain neuron names in an automatic fashion (without the use of any extra annotations) by feeding a one-hot encoded vector corresponding to a neuron being activated 
to $W_{ a \rightarrow {\cal K}}$ and performing a similar process of top-1 retrieval (as the textual explanations) to obtain the corresponding `\emph{neuron name}'. 

Fig~\ref{fig:explanations} provides qualitative examples for named neurons and their activation maps. The green block shows instances where the unseen class images were correctly classified by $\mathtt{NET}_{S\cup U}$. Conversely, those in red correspond to errors. The columns correspond to the class-labels, images, Grad-CAM visualizations for the class, textual explanations in the attribute space and top-3 neuron names responsible for the target class and their corresponding activation maps. For instance, notice that in the second row, for the image -- correctly classified as a yellow-headed blackbird -- the visualizations for the class focus specifically at the union of attributes that comprise the class. In addition, the textual explanations also filter out these attributes based on the neuron-importance scores - \emph{has throat color yellow}, \emph{has wing color black}, etc. In addition, when we focus on the individual neurons with relatively higher importance we see that individual neurons focus on the visual regions characterized by their assigned `names'. This shows that our neuron names are indeed representative of the concepts learned by the network and are well grounded in the image. 

Consider the misclassified examples (rows 7 and 8). 
Looking at the regions in the image corresponding to the intersection of the attributes in the textual explanations for the ground truth as well as the predicted class, we can see that the network was unable to focus on the primary discriminative attributes. 
Similarly, the neuron names and corresponding activations have a mismatch with the predicted class with the activation maps focusing on a `yellowish' area rather than a visual region corresponding to a fine-grained attribute. \\[-15pt]

\section{Conclusion \\[-15pt]}

\par \noindent 
To summarize, we propose an approach we refer to as Neuron Importance-aware Weight Transfer (NIWT), that learns to map domain knowledge about novel classes directly to classifier weights by grounding it into the importance of network neurons.  Our weight optimization approach on this grounding results in classifiers for unseen classes which outperform existing approaches on the  generalized zero-shot learning benchmark. We further demonstrate that this grounding between language and neurons can also be learned in reverse, linking neurons to human interpretable semantic concepts, providing visual and textual explanations. \\[-10pt]

\par \noindent
\textbf{Acknowledgements.} We thank Yash Goyal and Nirbhay Modhe for help with figures; Peter Vajda and Manohar Paluri for helpful discussions.
This work was supported in part by NSF, AFRL, DARPA, Siemens, Google, Amazon, ONR YIPs and ONR Grants N00014-16-1-\{2713,2793\}. 
The views and conclusions contained herein are those of the authors and should not be interpreted as necessarily representing the official policies or endorsements, either expressed or implied, of the U.S. Government, or any sponsor.

\bibliographystyle{splncs04}
\bibliography{strings,main}

\end{document}